\documentclass[sigconf]{acmart}

\usepackage{balance}
  
\usepackage[utf8]{inputenc} 
\usepackage{url}            
\usepackage{booktabs}       
\usepackage{nicefrac}       
\usepackage{microtype}      

\usepackage{algorithm,algpseudocode}
\usepackage{amsmath}
\usepackage{amsfonts}
\usepackage{subcaption}
\usepackage{bm}

\usepackage{graphicx}
\usepackage[american]{babel}
\usepackage{hyperref}
\usepackage{caption}
\usepackage{multirow}
\usepackage{enumitem}

\def\0{{\bf 0}}
\def\1{{\bf 1}}

\def\BibTeX{{\rm B\kern-.05em{\sc i\kern-.025em b}\kern-.08emT\kern-.1667em\lower.7ex\hbox{E}\kern-.125emX}}

\copyrightyear{2020} 
\acmYear{2020} 
\setcopyright{acmcopyright}\acmConference[MM '20]{Proceedings of the 28th ACM International Conference on Multimedia}{October 12--16, 2020}{Seattle, WA, USA}
\acmBooktitle{Proceedings of the 28th ACM International Conference on Multimedia (MM '20), October 12--16, 2020, Seattle, WA, USA}
\acmPrice{15.00}
\acmDOI{10.1145/3394171.3413561}
\acmISBN{978-1-4503-7988-5/20/10}

\begin{document}

\fancyhead{}

\title{F2GAN: Fusing-and-Filling GAN for Few-shot Image Generation}



\settopmatter{authorsperrow=3}
\author{Yan Hong}
\affiliation{%
  \institution{Shanghai Jiao Tong University}
}
\email{yanhong.sjtu@gmail.com}

\author{Li Niu}
\authornote{Corresponding author.}
\affiliation{%
  \institution{Shanghai Jiao Tong University}}
\email{ustcnewly@sjtu.edu.cn}

\author{Jianfu Zhang}
\affiliation{%
  \institution{Shanghai Jiao Tong University}
}
\email{c.sis@sjtu.edu.cn}

\author{Weijie Zhao}
\affiliation{%
  \institution{Versa-AI}
}
\email{weijie.zhao@versa-ai.com}

\author{Chen Fu}
\affiliation{%
  \institution{Versa-AI}
}
\email{chen.fu@versa-ai.com}

\author{Liqing Zhang}
\affiliation{\institution{Shanghai Jiao Tong University}}
\email{zhang-lq@cs.sjtu.edu.cn}

%
\renewcommand{\shortauthors}{Hong, et al.}

%

\begin{abstract}
  In order to generate images for a given category, existing deep generative models generally rely on abundant training images. However, extensive data acquisition is expensive and fast learning ability from limited data is necessarily required in real-world applications. Also, these existing methods are not well-suited for fast adaptation to a new category.
  Few-shot image generation, aiming to generate images from only a few images for a new category, has attracted some research interest. In this paper, we propose a Fusing-and-Filling Generative Adversarial Network (F2GAN) to generate realistic and diverse images for a new category with only a few images. In our F2GAN, a fusion generator is designed to fuse the high-level features of conditional images with random interpolation coefficients, and then fills in attended low-level details with non-local attention module to produce a new image. Moreover, our discriminator can ensure the diversity of generated images by a mode seeking loss and an interpolation regression loss. Extensive experiments on five datasets demonstrate the effectiveness of our proposed method for few-shot image generation.
\end{abstract}

%
%



%

%

\maketitle

\section{Introduction}
Deep generative models
, mainly including Variational Auto-Encoder (VAE) based methods~\cite{vae} and Generative Adversarial Network (GAN) based methods~\cite{goodfellow2014generative}, draw extensive attention from the artificial intelligence community. Despite the advances achieved in current GAN-based methods~\cite{cyclegan,stargan1, stargan2,stylegan1, stylegan2,mm2,DoveNet2020,GAIN2019}, the remaining bottlenecks in deep generative models are the necessity of amounts of training data and the difficulties with fast adaptation to a new category~\cite{clouatre2019figr,bartunov2018few,liang2020dawson}, especially for those newly emerging categories or long-tail categories. Therefore, it is necessary to consider how to generate images for a new category with only a few images. This task is referred to as few-shot image generation~\cite{clouatre2019figr,hong2020matchinggan}, which can benefit a wide range of downstream category-aware tasks like few-shot classification~\cite{vinyals2016matching,sung2018learning}. 

\begin{figure}
\begin{center}
\includegraphics[scale=0.32]{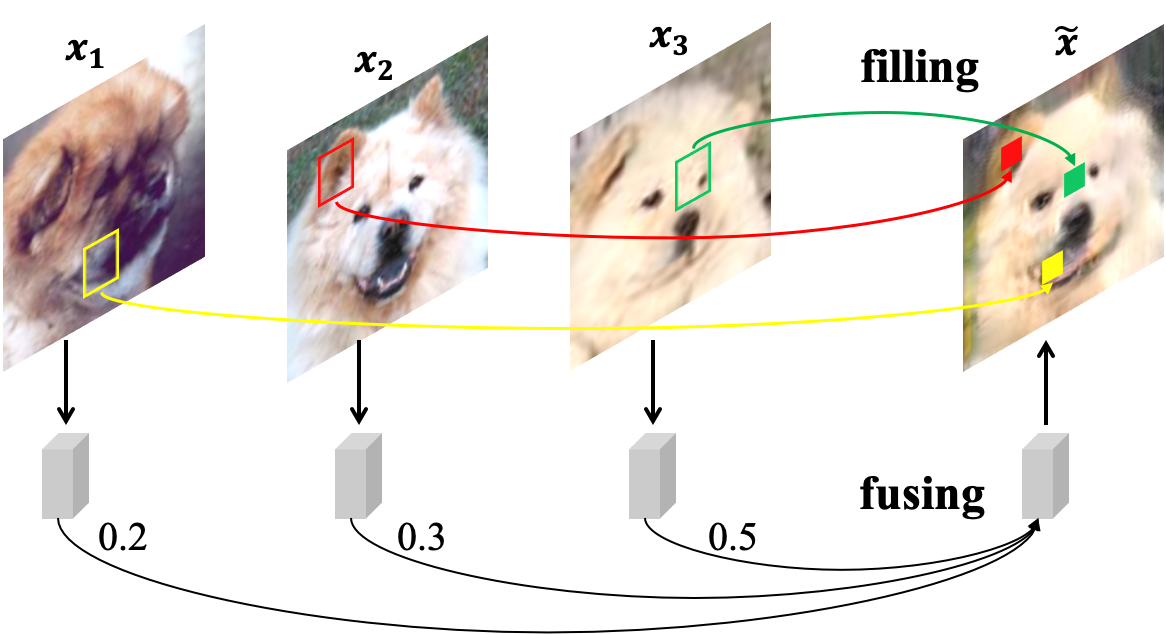}
\end{center}
\caption{Illustration of fusing three conditional images $\textbf{x}_1, \textbf{x}_2, \textbf{x}_3$ with interpolation coefficients $[ \textbf{0.2},  \textbf{0.3},  \textbf{0.5}]$ in our proposed F2GAN. The high-level features of conditional images are fused with interpolation coefficients and the details (\emph{e.g.}, color dots representing query locations) of the generated image are filled by using relevant low-level features (\emph{e.g.}, color boxes corresponding to query locations) from conditional images.  Best viewed in color.}
\label{fig:attention_explain} 
\end{figure}

In the few-shot image generation task, the model is trained on seen categories with sufficient labeled training images. Then, given only a few training images from a new unseen category, the learnt model is expected to produce more diverse and realistic images for this unseen category. In some previous few-shot image generation methods~\cite{vinyals2016matching,hong2020matchinggan}, the model is trained on seen categories in an episode-based manner~\cite{vinyals2016matching}, in which a small number (\emph{e.g.}, 1, 3, 5) of images from one seen category are provided in each training episode~\cite{vinyals2016matching} to generate new images. The input images used in each training episode are called conditional images. After training, the learnt model can generate new images by using a few conditional images from each unseen category.

To the best of our knowledge, there are quite few works on few-shot image generation. Among them, DAGAN~\cite{antoniou2017data} is a special case, \emph{i.e.}, one-shot image generation, which injects random noise into the generator to produce a slightly different image from the same category. However, this method is conditioned on only one image and fails to fuse the information of multiple images from the same category. More recent few-shot image generation methods can be divided into optimization-based methods and metric-based methods. Particularly, optimization-based FIGR~\cite{clouatre2019figr} (\emph{resp.}, DAWSON~\cite{liang2020dawson}) adopted a similar idea to Reptile~\cite{nichol2018first} (\emph{resp.},  MAML~\cite{finn2017model}), by initializing a generator with images from seen categories and fine-tuning the trained model with images from each unseen category. 
Metric-based method GMN~\cite{bartunov2018few} (\emph{resp.}, MatchingGAN~\cite{hong2020matchinggan}) is inspired by matching network~\cite{vinyals2016matching} and combines matching procedure with VAE (\emph{resp.}, GAN). However, FIGR, DAWSON, and GMN can hardly produce sharp and realistic images. MatchingGAN performs better, but has difficulty in fusing complex natural images.

\begin{figure*}
\begin{center}
\includegraphics[scale=0.25]{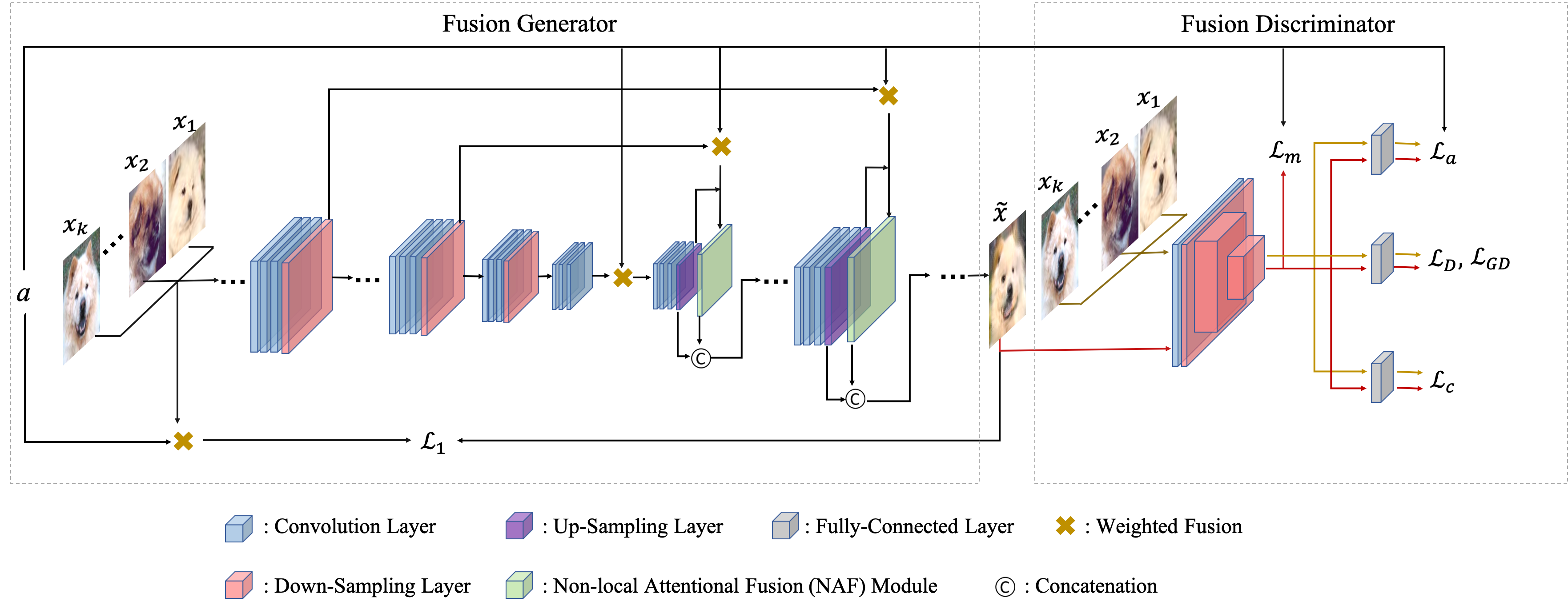}
\end{center}
\caption{The framework of our method which consists of a fusion generator and a fusion discriminator. $\tilde{ \textbf{x}}$ is generated based on the random interpolation coefficients $ \textbf{a}$ and $ \textbf{K}$ conditional images $\{ \textbf{x}_k|_{k=1}^K\}$. Due to space limitation, we only draw three encoder blocks and two decoder blocks. Best viewed in color.}
\label{fig:framework} 
\end{figure*}

In this paper, we follow the idea in \cite{hong2020matchinggan} by fusing conditional images, and propose a novel fusing-and-filling GAN (F2GAN) to enhance the fusion ability. The high-level idea is fusing the high-level features of conditional images and filling in the details of generated image with relevant low-level features of conditional images, which is depicted in Figure~\ref{fig:attention_explain}. In detail, our method contains a fusion generator and a fusion discriminator as shown in Figure~\ref{fig:framework}. Our generator is built upon U-Net structure with skip connections~\cite{ronneberger2015u} between the encoder and the decoder. A well-known fact is that in a CNN encoder, shallow blocks encode low-level information at high spatial resolution while deep blocks encode high-level information at low spatial resolution. We interpolate the high-level bottleneck features (the feature vector between encoder and decoder) of multiple conditional images with random interpolation coefficients. Then, the fused high-level feature is upsampled through the decoder to produce a new image. In each upsampling stage, we  borrow missing details from the skip-connected shallow encoder block by using our Non-local Attentional Fusion (NAF) module. Precisely, NAF module searches the outputs from shallow encoder blocks of conditional images in a global range, to attend the information of interest for each location in the generated image. 

In the fusion discriminator, we employ typical adversarial loss and classification loss to enforce the generated images to be close to real images and from the same category of conditional images. To ensure the diversity of generated images, we additionally employ a mode seeking loss and an interpolation regression loss, both of which are related to interpolation coefficients. Specifically, we use a variant of mode seeking loss~\cite{mao2019mode} to prevent the images generated based on different interpolation coefficients from collapsing to a few modes. Moreover, we propose a novel interpolation regression loss by regressing the interpolation coefficients based on the features of conditional images and generated image, which means that each generated image can recognize its corresponding interpolation coefficients. In the training phase, we train our F2GAN based on the images from seen categories. In the testing phase, conditioned on a few images from each unseen category, we can randomly sample interpolation coefficients to generate diverse images for this unseen category.

Our contributions can be summarized as follows: 1) we design a new few-shot image generation method F2GAN, by fusing high-level features and filling in low-level details; 2) Technically, we propose a novel non-local attentional fusion module in the generator and a novel interpolation regression loss in the discriminator; 3) Comprehensive experiments on five real datasets demonstrate the effectiveness of our proposed method.

\section{Related Work}
\label{sec:related}

\noindent\textbf{Data Augmentation:}
Data augmentation~\cite{Krizhevsky2012ImageNet} targets at augmenting the training set with new samples. Traditional data augmentation techniques (\emph{e.g.}, crop, rotation, color jittering) can only produce limited diversity. Some more advanced augmentation techniques~\cite{zhang2017mixup,yun2019cutmix} are proposed, but they fail to produce realistic images. 
In contrast, deep generative models can exploit the distribution of training data to generate more diverse and realistic samples for feature augmentation~\cite{schwartz2018delta,mm1} and image augmentation~\cite{antoniou2017data}. Our method belongs to image augmentation and can produce more images to augment the training set.

\noindent\textbf{Generative Adversarial Network:}
Generative Adversarial Network (GAN)~\cite{goodfellow2014generative,xu2019learning} is a powerful generative model based on adversarial learning. In the early stage, unconditional GANs~\cite{miyato2018spectral} generated images with random vectors by learning the distribution of training images. Then, GANs conditioned on a single image~\cite{miyato2018cgans,antoniou2017data} were proposed to transform the conditional image to a target image. Recently, a few conditional GANs attempted to accomplish more challenging tasks conditioned on more than one image, such as few-shot image translation~\cite{liu2019few} and few-shot image generation~\cite{clouatre2019figr,bartunov2018few}. 
In this paper, we focus on few-shot image generation, which will be detailed next.

\noindent\textbf{Few-shot Image Generation}
Few-shot generation is a challenging problem which can generate new images with a few conditional images. Early few-shot image generation works are limited to certain application scenario. For example, Bayesian learning and reasoning were applied in ~\cite{lake2011one,rezende2016one} to learn simple concepts like pen stroke and combine the concepts hierarchically to generate new images. 
More recently, FIGR~\cite{clouatre2019figr} was proposed to combine adversarial learning with optimization-based few-shot learning method Reptile~\cite{nichol2018first} to generate new images. Similar to FIGR~\cite{clouatre2019figr}, DAWSON~\cite{liang2020dawson}  applied meta-learning MAML algorithms~\cite{finn2017model} to GAN-based generative models to achieve domain adaptation between seen categories and unseen categories. Metric-based few-shot learning method Matching Network~\cite{vinyals2016matching} was combined with Variational Auto-Encoder~\cite{Pu2016Variational} in GMN~\cite{bartunov2018few} to generate new images without finetuning in the test phase. MatchingGAN~\cite{hong2020matchinggan} attempted to use learned metric to generate images based on a single or a few conditional images. In this work, we propose a new solution for few-shot image generation, which can generate more diverse and realistic images.

\noindent\textbf{Attention Mechanism:}
Attention module aims to localize the regions of interest. Abundant attention mechanisms like spatial attention~\cite{xu2016ask}, channel attention~\cite{chen2017sca}, and full attention~\cite{wang2018mancs} have been developed. Here, we discuss two works most related to our method. The method in~\cite{lathuiliere2019attention} employs local attention mechanism to select relevant information from multi-source human images for human image generation, but it fails to capture long-range relevance.  Inspired by non-local attention~\cite{zhang2019self,wang2018non}, we develop a novel non-local attentional fusion (NAF) module for few-shot image generation.

\section{Our Method}
Given a few conditional images $\mathcal{X}_S=\{\bm{x}_k|_{k=1}^K\}$ from the same category ($K$ is the number of conditional images) and random interpolation coefficients $\bm{a}=[a^1,\ldots,a^K]$, our model targets at generating a new image from the same category. We fuse the high-level bottleneck features of conditional images $\{\bm{x}_k|_{k=1}^K\}$  with interpolation coefficients $\bm{a}$, and fill in the low-level details specified by Non-local Attentional Fusion (NAF) module during upsampling to generate a new image $\tilde{\bm{x}}$. 

We split all categories into seen categories  $\mathcal{C}^{s}$ and unseen categories $\mathcal{C}^{u}$, where $\mathcal{C}^{s} \cap \mathcal{C}^{u}=\emptyset$. In the training phase, our model is trained with images from seen categories $\mathcal{C}^{s}$ to learn a mapping, which translates a few conditional images $\mathcal{X}_S$ of a seen category to a new image belonging to the same category. In the testing phase, a few conditional images from an unseen category in $\mathcal{C}^{u}$ together with random interpolation coefficients $\bm{a}$ are fed into the trained model to generate new diverse images for this unseen category. As illustrated in Figure~\ref{fig:framework}, our model consists of a fusion generator and a fusion discriminator, which will be detailed next.

\begin{figure}
\begin{center}
\includegraphics[scale=0.36]{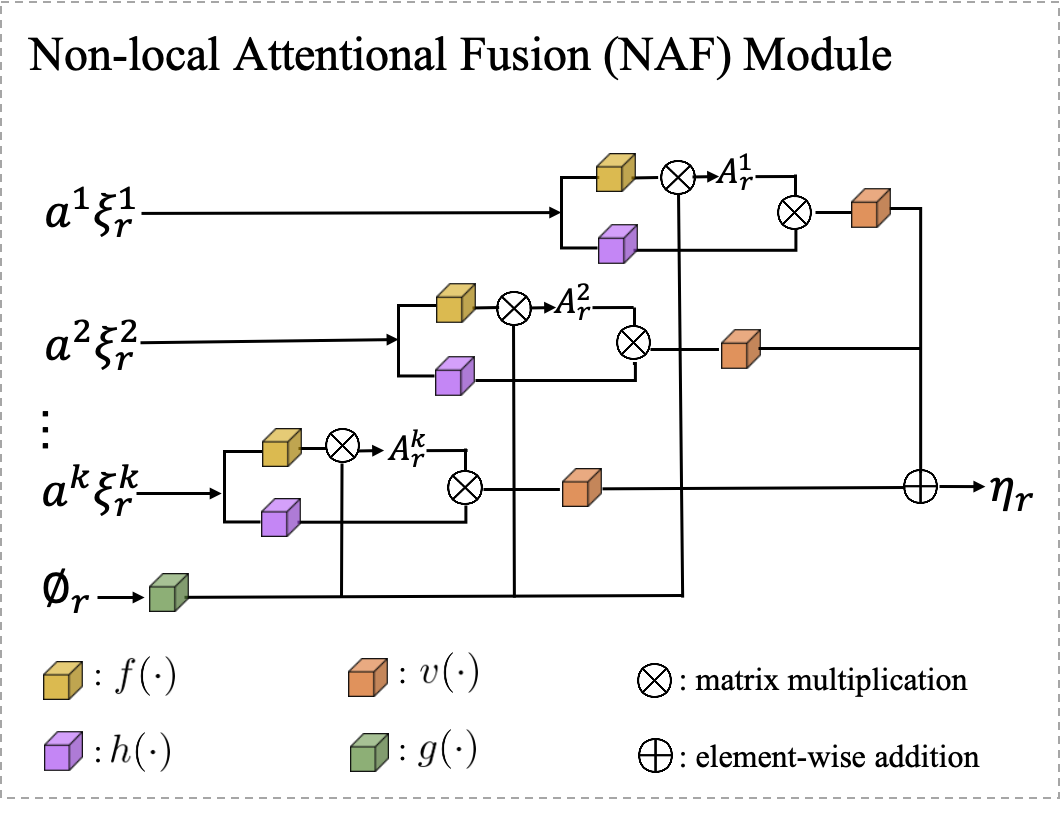}
\end{center}
\caption{The architecture of our Non-local Attentional Fusion (NAF) module. $\bm{\xi}_r^k$ is the feature of $\textbf{x}_k$ from the $r$-th encoder block, $\bm{\phi}_r$ is the output from the $r$-th decoder block, and $\bm{\eta}_r$ is the output of NAF.}

\label{fig:attention} 
\end{figure}

\subsection{Fusion Generator}
Our fusion generator $G$ adopts an encoder-decoder structure~\cite{antoniou2017data} which is a combination of U-Net~\cite{ronneberger2015u} and ResNet~\cite{he2016deep}. Specifically, there are in total $11$  residual blocks ($5$ encoder blocks, $5$ decoder blocks, and $1$ intermediate block), in which each encoder (\emph{resp.}, decoder) block contains $3$ convolutional layers with leaky ReLU and batch normalization followed by one downsampling (\emph{resp.}, upsampling) layer, while the intermediate block contains $3$ convolutional layers with leaky ReLU and batch normalization. The detailed architecture can be found in Supplementary. The encoder (\emph{resp.}, decoder) blocks progressively decrease (\emph{resp.}, increase) the spatial resolution. For ease of description, the encoder (\emph{resp.}, decoder) blocks from shallow to deep are indexed from $4$ (\emph{resp.}, $1$) to $0$ (\emph{resp.}, $5$). We use $\bm{{\psi}}^k$ to denote the bottleneck feature of $\bm{x}_k$ from the intermediate block. Besides, we add $3$ skip connections between the encoder and the decoder. For $r=1,2,3$, the $r$-th skip connection directs the output from the $r$-th encoder block to the output from the $r$-th decoder block.
Then, we use $\bm{{\xi}}_r^k \in  \mathcal{R}^{W_r \times H_r \times C_r}$ to denote the output feature of conditional image $\bm{x}_k$ from the $r$-th encoder block, and $\bm{{\phi}}_r \in  \mathcal{R}^{W_r \times H_r \times C'_r}$ to denote the output feature from the $r$-th decoder block, where $C_r$ and $C'_r$ are the number of channels in the $r$-th encoder and decoder respectively. 

To fuse the bottleneck features of conditional images $\mathcal{X}_S$, we randomly sample interpolation coefficients $\bm{a}=[a^1,\ldots,a^K]$, which satisfy $a^k\geq 0$ and $\sum_{k=1}^K a^k=1$, leading to the fused bottleneck feature $\bm{{\eta}}_0 =\sum_{k=1}^{K} a^{k} \bm{{\psi}}^k$.
Since the spatial size of bottleneck feature is very small (\emph{e.g.}, $4\times 4$), the spatial misalignment issue can be ignored and high-level semantic information of conditional images is fused. 
Then, the fused bottleneck feature is upsampled through decoder blocks. During upsampling in each decoder block, lots of details are missing and need to be filled in.
We borrow the low-level detailed information from the output features of its skip-connected encoder block. Furthermore, we insert a Non-local Attentional Fusion (NAF) module into the skip connection to attend relevant detailed information, as shown in Figure~\ref{fig:framework}. For the $r$-th skip connection, NAF module takes $a^k\bm{\xi}_r^k$ and $\bm{\phi}_r$ as input and outputs $\bm{{\eta}}_r = \textnormal{NAF}(\{a^k\bm{\xi}_r^k|_{k=1}^K\}, \bm{\phi}_r)$.
Then, $\bm{{\eta}}_r$ concatenated with $\bm{\phi}_r$, that is, $\hat{\bm{{\phi}}}_{r} = [\bm{{\eta}}_r, \bm{{\phi}}_r]$, is taken as the input to the $(r+1)$-th decoder block.

Our attention-enhanced fusion strategy is a little similar to~\cite{lathuiliere2019attention}. However, for each spatial location on $\bm{\phi}_r$, the attention module in~\cite{lathuiliere2019attention} only attends exactly the same location on $\bm{\xi}_r^k$, which will hinder attending relevant information if the conditional images are not strictly aligned. For example, for category ``dog face", the dog eyes may appear at different locations in different conditional images which have different face poses. Inspired by non-local attention~\cite{zhang2019self,wang2018non}, for each spatial location on $\bm{\phi}_r$, we search relevant information in a global range on $\bm{\xi}_r^k$. Specifically, our proposed Non-local Attentional Fusion (NAF) module calculates an attention map $\bm{A}_r^k$ based on $\bm{{\phi}}_r$ and $a^k\bm{{\xi}}_r^k$, in which each entry ${A}_r^k(i,j)$ represents the attention score between the $i$-th location on $\bm{{\phi}}_r$ and the $j$-th location on $a^k\bm{{\xi}}_r^k$. Therefore, the design philosophy  and technical details of our NAF module are considerably different from those in~\cite{lathuiliere2019attention}. 

The architecture of NAF module is shown in Figure~\ref{fig:attention}. First, $a^k\bm{{\xi}}_r^k$ and $\bm{\phi}_r$ are projected to a common space by $f(\cdot)$ and $g(\cdot)$ respectively, where $f(\cdot)$  and $g(\cdot)$ are $1 \times 1 \times \frac{C_r}{8}$ convolutional layer with spectral normalization~\cite{miyato2018spectral}. For ease of calculation, we reshape $f(a^k\bm{{\xi}}_r^k) \in  \mathcal{R}^{W_r \times H_r \times \frac{C_r}{8}} $ (\emph{resp.}, $g(\bm{{\phi}}_r) \in  \mathcal{R}^{W_r \times H_r \times \frac{C_r}{8}}$) into $\bar{f}(a^k\bm{{\xi}}_r^k) \in  \mathcal{R}^{N_r \times \frac{C_r}{8}}$ (\emph{resp.}, $\bar{g}(\bm{{\phi}}_r) \in  \mathcal{R}^{N_r \times \frac{C_r}{8}}$), in which $N_r=W_r \times H_r$. Then, we can calculate the attention map between $\bm{{\phi}}_r$ and $a^k\bm{{\xi}}_r^k$:
\begin{equation}\label{eqn:attention_map}
\begin{aligned}
\bm{A}_r^k = softmax\left(\bar{g}(\bm{{\phi}}_r)\bar{f}(a^k\bm{{\xi}}_r^k)^{T} \right).
\end{aligned}
\end{equation}
With obtained attention map $\bm{A}_r^k$, we attend information from $a^k\bm{{\xi}}_r^k$  and achieve the attended feature map $\bm{{\eta}}_r$:
\begin{equation}
\begin{aligned}
\bm{{\eta}}_r = \sum_{k=1}^{K} v\left(\bm{A}_r^k \bar{h}(a^k\bm{{\xi}}_r^k)\right),
\end{aligned}
\end{equation}
where $\bar{h}(\cdot)$ means $1 \times 1  \times  \frac{C_r}{8} $ convolutional layer followed by reshaping to $\mathcal{R}^{N_r \times \frac{C_r}{8}}$, similar to $\bar{f}(\cdot)$ and $\bar{g}(\cdot)$ in (\ref{eqn:attention_map}). $v(\cdot)$ reshapes the feature map back to $\mathcal{R}^{W_r \times H_r \times \frac{C_r}{8}}$ and then performs $1 \times 1  \times  \frac{C_r}{8} $ convolution.
 

As the  shallow (\emph{resp.}, deep) encoder block contains the low-level (\emph{resp.}, high-level) information, our generated images can fuse multi-level information of conditional images coherently. Finally, the generated image can be represented by $\tilde{\bm{x}} = G(\bm{a}, \mathcal{X}_S)$.

Following \cite{hong2020matchinggan}, we adopt a weighted reconstruction loss to constrain the generated image:
\begin{equation} \label{eqn:loss_reconstruction}
\begin{aligned}
\mathcal{L}_1 = \sum_{k=1}^{K} a^k || \bm{x}_k - \tilde{\bm{x}}||_1.
\end{aligned}
\end{equation}
Intuitively, the generated image should bear more resemblance to the conditional image with larger interpolation coefficient.

\begin{figure*}
\begin{center}
\includegraphics[scale=0.3]{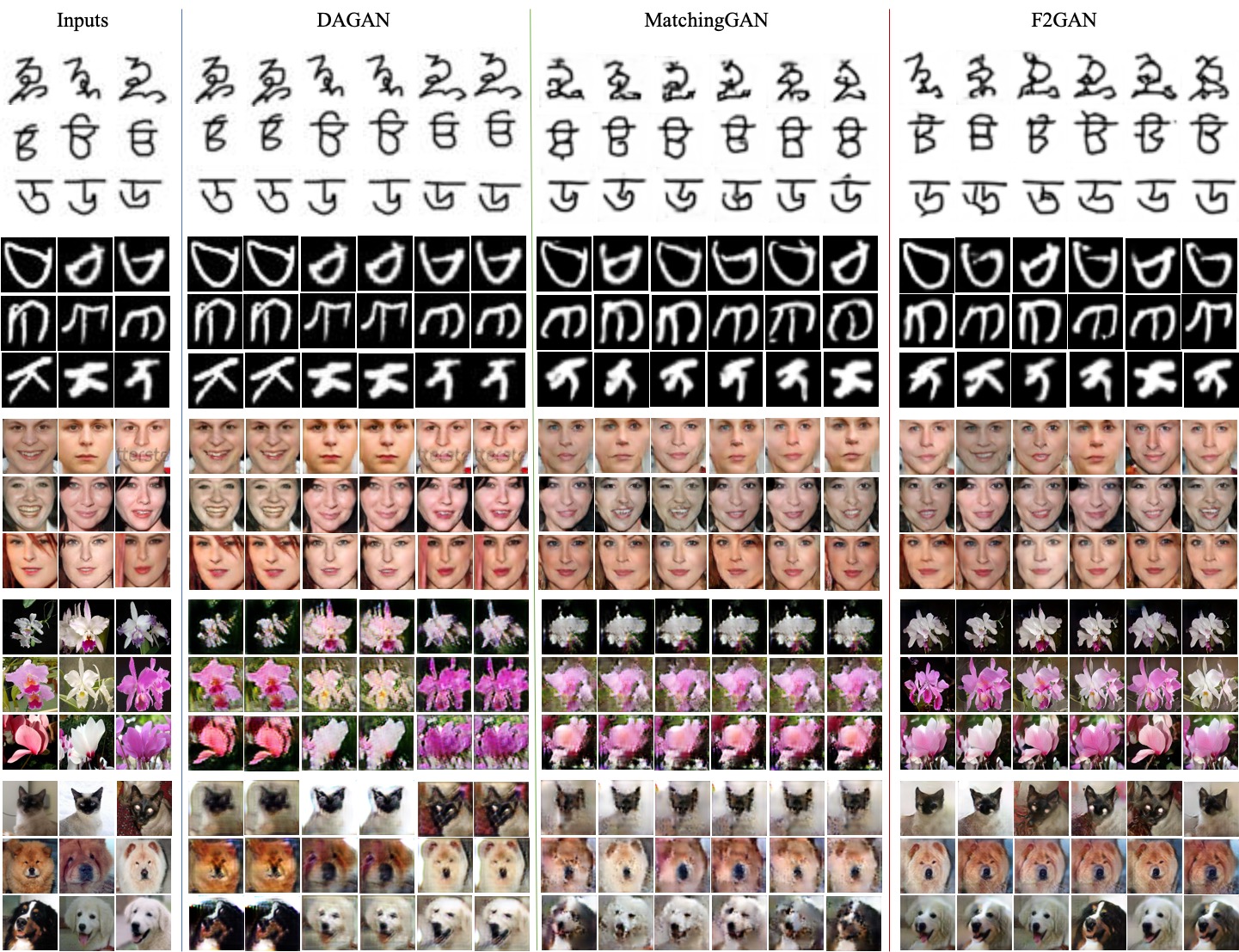}
\end{center}
\caption{Images generated by DAGAN, MatchingGAN, and our F2GAN ($ \textbf{K=3}$) on five datasets (from top to bottom: Omniglot, EMNIST, VGGFace, Flowers, and Animals Faces). The conditional images are in the left three columns.}
\label{fig:visualization} 
\end{figure*}

\subsection{Fusion Discriminator}
The network structure of our discriminator is analogous to that in~\cite{liu2019few}, which consists of one convolutional layer followed by five residual blocks~\cite{mescheder2018training}. The detailed architecture can be found in Supplementary. Differently, we use one fully connected (fc) layer with $1$ output following average pooling layer to obtain the discriminator score. We treat $K$ conditional images $\{\bm{x}_k|_{k=1}^K\}$ as real images and the generated image $\tilde{\bm{x}}$ as fake image. In detail, the average score $\mathrm{D}(\bm{x})$ for $K$ conditional images and the score $\mathrm{D}(\tilde{\bm{x}})$ for generated image $\tilde{\bm{x}}$ are calculated for adversarial learning. To stabilize training process, we use hinge adversarial loss in~\cite{miyato2018cgans}. To be exact, the goal of discriminator $\mathrm{D}$ is minimizing $\mathcal{L}_D$ while the goal of generator is minimizing $\mathcal{L}_{GD}$:
\begin{eqnarray}
\!\!\!\!\!\!\!\!&&\mathcal{L}_D = \mathbb{E}_{\tilde{\bm{x}}} [\max (0,1+\mathrm{D}(\tilde{\bm{x}})]  +  \mathbb{E}_{\bm{x}_k}  [\max (0,1-\mathrm{D}({\bm{x}}))], \nonumber\\
\!\!\!\!\!\!\!\!&&\mathcal{L}_{GD} = - \mathbb{E}_{\tilde{\bm{x}}} [\mathrm{D}(\tilde{\bm{x}})].
\end{eqnarray}

Analogous to ACGAN~\cite{odena2017conditional}, we apply a classifier with cross-entropy loss to classify the real images and the generated images into the corresponding seen categories.
Specifically, the last fc layer of the discriminator $D$ is replaced by another fc layer with the output dimension being the number of seen categories:
\begin{equation}
\begin{aligned}\label{eqn:loss_classification}
\mathcal{L}_{c} = -\log p(c(\bm{x})|\bm{x}),
\end{aligned}
\end{equation}
where $c(\bm{x})$ is the ground-truth category of $\bm{x}$. We minimize $\mathcal{L}^{D}_{c} = -\sum_{k=1}^K\log p(c(\bm{x}_k)|\bm{x}_k)$ for $K$ conditional images $\{\bm{x}_k|_{k=1}^K\}$ when training the discriminator. We update the generator by minimizing $\mathcal{L}^{G}_{c}=-\log p(c(\tilde{\bm{x}})|\tilde{\bm{x}})$, since we expect the generated image $\tilde{\bm{x}}$ to be classified as the same category of conditional images.


By varying interpolation coefficients  $\bm{a}$, we expect to generate diverse images, but one common problem for GAN is mode collapse~\cite{mao2019mode}, which means that the generated images may collapse into a few modes.
In our fusion generator, when sampling two different interpolation coefficients $\bm{a}_1$ and $\bm{a}_2$, the generated images $G(\bm{a}_1,\mathcal{X}_S)$ and $G(\bm{a}_2,\mathcal{X}_S)$ are likely to collapse into the same mode. To guarantee the diversity of generated images, we use two strategies to mitigate mode collapse, one is a variant of mode seeking loss~\cite{mao2019mode} to seek for more modes, the other is establishing bijection between the generated image $\tilde{\bm{x}}$ and its corresponding interpolation coefficient $\bm{a}$. The mode seeking loss in~\cite{mao2019mode} was originally used to produce diverse images when using different latent codes. Here, we slightly twist the mode seeking loss to produce diverse images when using different interpolation coefficients. Specifically,
we remove the last fc layer of $D$ and use the remaining feature extractor $\hat{D}$ to extract the features of generated images with different interpolation coefficients. Then, we maximize the ratio of the distance between $\hat{D}(G(\bm{a}_1,\mathcal{X}_S))$ and $\hat{D}(G(\bm{a}_2,\mathcal{X}_S))$ over the distance between $\bm{a}_1$ and $\bm{a}_2$, yielding the following mode seeking loss:
\begin{equation} \label{eqn:loss_mode_seeking}
\begin{aligned}
\mathcal{L}_{m} = \frac {|| \hat{D}(G(\bm{a}_1,\mathcal{X}_S)) -  \hat{D}(G(\bm{a}_2,\mathcal{X}_S))||_1} {|| \bm{a}_1 - \bm{a}_2||_1}.
\end{aligned}
\end{equation}

To further ensure the diversity of generated images, the bijection between the generated image $\tilde{\bm{x}}$ and its corresponding interpolation coefficient $\bm{a}$ is established by a novel interpolation regression loss, which regresses the interpolation coefficient $\bm{a}$ based on the features of conditional images $\hat{D}(\bm{x}_k)$ and generated image $\hat{D}(\tilde{\bm{x}})$. Note that the feature extractor $\hat{D}$ is the same as in (\ref{eqn:loss_mode_seeking}).  Specifically, we apply a fully-connected (fc) layer $E$ to the concatenated feature $[\hat{D}(\bm{x}_k),\hat{D}(\tilde{\bm{x}})]$, and obtain the similarity score $s_k$ between $\bm{x}_k$ and $\tilde{\bm{x}}$: $s_k =  E([\hat{D}(\bm{x}_k), \hat{D}(\tilde{\bm{x}})])$.
Then, we apply softmax layer to $\bm{s}=[s_1,\ldots,s_K]$ to obtain the predicted interpolation coefficients $\tilde{\bm{a}} = softmax(\bm{s})$, which are enforced to match the ground-truth $\bm{a}$:
\begin{equation} \label{eqn:loss_interpolation}
\begin{aligned}
\mathcal{L}_{a} = ||\tilde{\bm{a}}  - \bm{a} ||_2.
\end{aligned}
\end{equation}
By recognizing the interpolation coefficient based on the generated image and conditional images, we actually establish a bijection between the generated image and interpolation coefficient, which discourages two different interpolation coefficients from generating the same image. 

\subsection{Optimization}
The overall loss function to be minimized is as follows, 
\begin{equation}
\begin{aligned}
\mathcal{L} =  \mathcal{L}_D +  \mathcal{L}_{GD}+ \lambda_1 \mathcal{L}_{1} + \mathcal{L}_{c} - \lambda_m \mathcal{L}_{m} + \lambda_a \mathcal{L}_{a},\label{optimization}
\end{aligned}
\end{equation}
in which $\lambda_1$, $\lambda_m$, and $\lambda_a$ are trade-off parameters. In the framework of adversarial learning, fusion generator and fusion discriminator are optimized by related loss terms in an alternating manner. In particular, the fusion discriminator is optimized by minimizing $\mathcal{L}_D$ and $\mathcal{L}^{D}_c$, while the fusion generator is optimized by minimizing $\mathcal{L}_{GD}$, $\mathcal{L}_{1}$, $\mathcal{L}^{G}_{c}$, $-\mathcal{L}_{m}$, and $\mathcal{L}_{a}$, in which $\mathcal{L}^{D}_c$ and $\mathcal{L}^{G}_{c}$ are defined below (\ref{eqn:loss_classification}). 


\setlength{\tabcolsep}{4pt}
\begin{table*}[t]
  \caption{FID ($\downarrow$), IS ($\uparrow$) and LPIPS ($\uparrow$) of images generated by different methods for unseen categories on three datasets.} 
  \centering
  \begin{tabular}{lrrrrrrrrr}
      \toprule[0.8pt]
      \multirow{2}{*}{Method}&
      \multicolumn{3}{c}{VGGFace} & \multicolumn{3}{c}{Flowers} &\multicolumn{3}{c}{Animals Faces}\cr
       &FID ($\downarrow$)  & IS ($\uparrow$) & LPIPS($\uparrow$)  & FID ($\downarrow$)  & IS ($\uparrow$) & LPIPS ($\uparrow$) &FID ($\downarrow$)  & IS ($\uparrow$) & LPIPS ($\uparrow$) \cr
       \cmidrule(r){2-4}  \cmidrule(r){5-7}  \cmidrule(r){8-10}
    FIGR~\cite{clouatre2019figr} &139.83 &2.98 &0.0834 & 190.12&1.38 &0.0634 &211.54 &1.55 &0.0756\cr
    
    GMN~\cite{bartunov2018few}&136.21 &2.14 &0.0902 &200.11 &1.42 &0.0743 &220.45 &1.71 &0.0868  \cr
    DAWSON~\cite{liang2020dawson} &137.82 &2.56 & 0.0769 &  188.96& 1.25 &0.0583  & 208.68 &1.51 &0.0642 \cr
    DAGAN~\cite{antoniou2017data}& 128.34 & 4.12 & 0.0913& 151.21&2.18 &0.0812 &155.29 &3.32 &0.0892\cr
    MatchingGAN~\cite{hong2020matchinggan}& 118.62 & 6.16 & 0.1695&  143.35& 4.36&0.1627 & 148.52& 5.08& 0.1514\cr
    F2GAN &$\textbf{109.16}$  &$\textbf{8.85}$ & $\textbf{0.2125}$ &$\textbf{120.48}$ &$\textbf{6.58}$ &$\textbf{0.2172}$ &$\textbf{117.74}$ &$\textbf{7.66}$ &$\textbf{0.1831}$\cr
    \bottomrule[0.8pt]
    
  \end{tabular}
  \vspace{0.1mm}
  \label{tab:performance_metric}
\end{table*}

\section{Experiments}
\label{sec:experiments}

\setlength{\tabcolsep}{4pt}
\begin{table*}[t]
  \caption{Accuracy(\%) of different methods on three datasets in low-data setting.} 
  \centering
  \begin{tabular}{lrrrrrrrrr}
      \toprule[0.8pt]
      \multirow{2}{*}{Method}&
      \multicolumn{3}{c}{Omniglot } & \multicolumn{3}{c}{EMNIST} &\multicolumn{3}{c}{VGGFace}\cr
       &5-sample   & 10-sample & 15-sample  & 5-sample  & 10-sample & 15-sample &5-sample  & 10-sample &15-sample \cr
       \cmidrule(r){2-4}  \cmidrule(r){5-7}  \cmidrule(r){8-10}
    Standard &66.22  & 81.87 &83.31  & 83.64 & 88.64 & 91.14 & 8.82 & 20.29 & 39.12\cr
    Traditional &67.32  &82.28  & 83.95 & 84.62  & 89.63 & 92.07 & 9.12 &22.83  & 41.63 \cr
    
    FIGR~\cite{clouatre2019figr} & 69.23  & 83.12 & 84.89 & 85.91  & 90.08 & 92.18 & 6.12  &  18.84& 32.13 \cr
    GMN~\cite{bartunov2018few} &  67.74 & 84.19 &  85.12  & 84.12  & 91.21 & 92.09 & 5.23  & 15.61 &35.48\cr
    DAWSON~\cite{liang2020dawson} &68.56 &82.02 & 84.01 & 83.63 & 90.72 & 91.83 & 5.27 &16.92 &30.61 \cr
    DAGAN~\cite{antoniou2017data} & 88.81  &89.32 & 95.38  &87.45  & 94.18& 95.58 &19.23 &35.12 & 44.36\cr
    MatchingGAN~\cite{hong2020matchinggan} &89.03  &90.92 & 96.29  & 91.75 & 95.91 &96.29  &21.12  &40.95 & 50.12\cr
    F2GAN &$\textbf{91.93}$ &$\textbf{92.48}$ & $\textbf{97.12}$& $\textbf{93.18}$& $\textbf{97.01}$ &$\textbf{97.82}$ & $\textbf{24.76}$&$\textbf{43.21}$ & $\textbf{53.42}$\cr

    \bottomrule[0.8pt]
    
  \end{tabular}
  \vspace{0.1mm}
  \label{tab:performance_vallia_classifier}
\end{table*}

\setlength{\tabcolsep}{2pt}
\begin{table*}[t]
  \caption{Accuracy(\%) of different methods on three datasets in few-shot classification setting.} 
  \centering
  \begin{tabular}{lrrrrrr}
      \toprule[0.8pt]
      \multirow{2}{*}{Method}&\multicolumn{2}{c}{VGGFace}&\multicolumn{2}{c}{Flowers}&\multicolumn{2}{c}{Animals Faces}
      \cr & 5-way 5-shot &10-way 5-shot & 5-way 5-shot &10-way 5-shot & 5-way 5-shot &10-way 5-shot\cr
      \cmidrule(r){2-3}  \cmidrule(r){4-5}  \cmidrule(r){6-7}
    MatchingNets~\cite{vinyals2016matching} & 60.01 &48.67  & 67.98&56.12  &59.12  &50.12 \cr

    MAML~\cite{finn2017model} & 61.09&47.89   & 68.12&58.01  & 60.03  &49.89 \cr

    RelationNets~\cite{sung2018learning}& 62.89 & 54.12 &69.83&61.03  &67.51  & 58.12 \cr

    MTL~\cite{sun2019meta}&77.82  &68.95  &82.35 &74.24  &79.85  &70.91 \cr

    DN4~\cite{li2019revisiting}&78.13  &70.02  &83.62 &73.96 &81.13  &71.34 \cr
    
    MatchingNet-LFT~\cite{Hungfewshot} &77.64  &69.92  & 83.19 &74.32  &80.95  &71.62 \cr
    MatchingGAN~\cite{hong2020matchinggan} & 78.72 & 70.94 &82.76 & 74.09 & 80.36 &  70.89\cr

    F2GAN&$\textbf{79.85}$  &$\textbf{72.31}$  &$\textbf{84.92}$ &$\textbf{75.02}$  &$\textbf{82.69}$  &$\textbf{73.19}$ \cr
    \bottomrule[0.8pt]
    \end{tabular}
  \vspace{0.1mm}
  \label{tab:performance_fewshot_classifier}
\end{table*}


\subsection{Datasets and Implementation Details}
We conduct experiments on five real datasets including Omniglot \cite{Brenden2015One}, EMNIST~\cite{cohen2017emnist},  VGGFace~\cite{cao2018vggface2}, Flowers~\cite{nilsback2008automated}, and Animal Faces~\cite{deng2009imagenet}. For VGGFace (\emph{resp.}, Omniglot, EMNIST), following MatchingGAN \cite{hong2020matchinggan}, we randomly select $1802$ (\emph{resp.}, $1200$, $28$) categories from total $2395$ (\emph{resp.}, $1623$, $48$) categories as training seen categories and select $497$ (\emph{resp.}, $212$, $10$) categories from remaining categories as unseen testing categories. For Animal face and flower datasets, we use the seen/unseen split provided in~\cite{liu2019few}.  In Animal Faces, $117574$ animal faces from $149$ carnivorous animal categories are selected from ImageNet~\cite{deng2009imagenet}. All animal categories are split into $119$ seen categories for training and $30$ unseen categories for testing. For Flowers dataset with $8189$ images distributed in $102$ categories, there are $85$ training seen categories and $17$ testing unseen categories.

We set $\lambda_1=1$, $\lambda_m = 0.01$, and $\lambda_a = 1$ in (\ref{optimization}). We set the number of conditional images $K=3$ by balancing the benefit against the cost, because larger $K$ only brings slight improvement (see Supplementary). We use Adam optimizer with learning rate 0.0001 and train our model for $200$ epochs.

\subsection{Quantitative Evaluation of Generated Images} \label{sec:visualization}
We evaluate the quality of images generated by different methods on three datasets based on commonly used Inception Scores (IS)~\cite{xu2018empirical}, Fréchet Inception Distance (FID)~\cite{heusel2017gans}, and Learned Perceptual Image Patch Similarity (LPIPS)~\cite{zhang2018unreasonable}. The IS is positively correlated with visual quality of generated images. We fine-tune the ImageNet-pretrained Inception-V3 model~\cite{szegedy2016rethinking} with unseen categories to calculate the IS for generated images. The FID is designed for measuring similarities between two sets of images. We remove the last average pooling layer of the ImageNet-pretrained Inception-V3 model as the feature extractor. Based on the extracted features, we compute Fréchet Inception Distance between the generated images and the real images from the unseen categories. The LPIPS can be used to measure the average feature distance among the generated images. We compute the average of pairwise distances among generated images for each category, and then compute the average over all unseen categories as the final LPIPS score. The details of distance calculation can be found in \cite{zhang2018unreasonable}.

For our method, we train our model based on seen categories. Then, we use a random interpolation coefficient and $K=3$ conditional images from each unseen category to generate a new image for this unseen category. We can generate adequate images for each unseen category by repeating the above procedure.
Similarly, GMN~\cite{bartunov2018few}, FIGR~\cite{clouatre2019figr} and MatchingGAN~\cite{hong2020matchinggan} are trained in $1$-way $3$-shot setting based on seen categories, and the trained models are used to generate images for unseen categories. Different from the above methods, DAGAN~\cite{antoniou2017data} is conditioned on a single image, but we can use one conditional image each time to generate adequate images for unseen categories.

For each unseen category, we use each method to generate $128$ images based on sampled $30$ real images, and calculate FID, IS and LPIPS based on the generated images. The results of different methods are reported in Table \ref{tab:performance_metric}, from which we observe that our method achieves the highest IS, lowest FID, and highest LPIPS, demonstrating that our model could generate more diverse and realistic images compared with baseline methods.

We show some example images generated by our method on five datasets including simple concept datasets and relatively complex natural datasets in Figure~\ref{fig:visualization}. For comparison, we also show the images generated by DAGAN and MatchingGAN, which are competitive baselines as demonstrated in Table \ref{tab:performance_metric}. On concept datasets Omniglot and EMNIST, we can see that the images generated by DAGAN are closer to inputs with limited diversity, while MatchingGAN and F2GAN can both fuse features from conditional images to generate diverse images for simple concepts. On natural datasets VGGFace, Flowers, and Animals Faces, we observe that MatchingGAN can generate plausible images on VGGFace dataset because face images are well-aligned. However, the images generated by MatchingGAN are of low quality on Flowers and Animals Faces datasets. In contrast, the images generated by our method are more diverse and realistic than DAGAN and MatchingGAN, because the information of more than one conditional image are fused more coherently in our method. In Supplementary, we also visualize our generated results on FIGR-8 dataset, which is released and used in FIGR~\cite{clouatre2019figr}, as well as more visualization results on Flowers and Animals datasets.

\subsection{Visualization of Linear Interpolation}
To evaluate whether the space of generated images is densely populated, we perform linear interpolation based on two conditional images $\bm{x}_1$ and $\bm{x}_2$  for ease of visualization. In detail, for interpolation coefficients $\bm{a}=[a^1, a^2]$, we start from $[0.9, 0.1]$, and then gradually decrease (\emph{resp.}, increase) $a^1$ (\emph{resp.},  $a^2$) to $0.1$ (\emph{resp.}, $0.9$) with step size $0.1$. Because MatchingGAN also fuses conditional images with interpolation coefficients, we report the results of both MatchingGAN and our F2GAN in Figure~\ref{fig:interpolation}. Compared with MatchingGAN, our F2GAN can produce more diverse images with smoother transition between two conditional images. More results can be found in Supplementary.

\subsection{Low-data Classification}
\label{sec:vallia}
To further evaluate the quality of generated images, we use generated images to help downstream classification tasks in low-data setting in this section and few-shot setting in Section \ref{sec:few-shot}. For low-data classification on unseen categories, following MatchingGAN~\cite{hong2020matchinggan}, we randomly select a few (\emph{e.g.}, $5$, $10$, $15$) training images per unseen category while the remaining images in each unseen category are test images. Note that we have training and testing phases for the classification task, which are different from the training and testing phases of our F2GAN.
We initialize the ResNet$18$~\cite{he2016deep} backbone using the images of seen categories, and then train the classifier using the training images of unseen categories. Finally, the trained classifier is used to predict the test images of unseen categories. This setting is referred to as ``Standard" in Table~\ref{tab:performance_vallia_classifier}. 

Then, we use the generated images to augment the training set of unseen categories. For each few-shot generation method, we generate $512$ images for each unseen category based on the training set of unseen categories. Then, we train the ResNet$18$ classifier on the augmented training set (including both original training set and generated images) and apply the trained classifier to the test set of unseen categories. We also use traditional augmentation techniques (\emph{e.g.}, crop, rotation, color jittering) to augment the training set and report the results as ``Traditional" in Table~\ref{tab:performance_vallia_classifier}.

The results of different methods are listed in Table~\ref{tab:performance_vallia_classifier}. On Omniglot and EMNIST datasets, all methods outperform ``Standard" and ``Traditional", which demonstrates the benefit of deep augmentation methods. On VGGFace dataset, our F2GAN, MatchingGAN~\cite{hong2020matchinggan}, and DAGAN~\cite{antoniou2017data} outperform ``Standard",  while the other methods underperform ``Standard". One possible explanation is that the images generated by GMN and FIGR on VGGFace are of low quality, which harms the classifier. It can also be seen that our proposed F2GAN achieves significant improvement over baseline methods, which corroborates the high quality of our generated images.

\begin{figure}
\begin{center}
\includegraphics[scale=0.3]{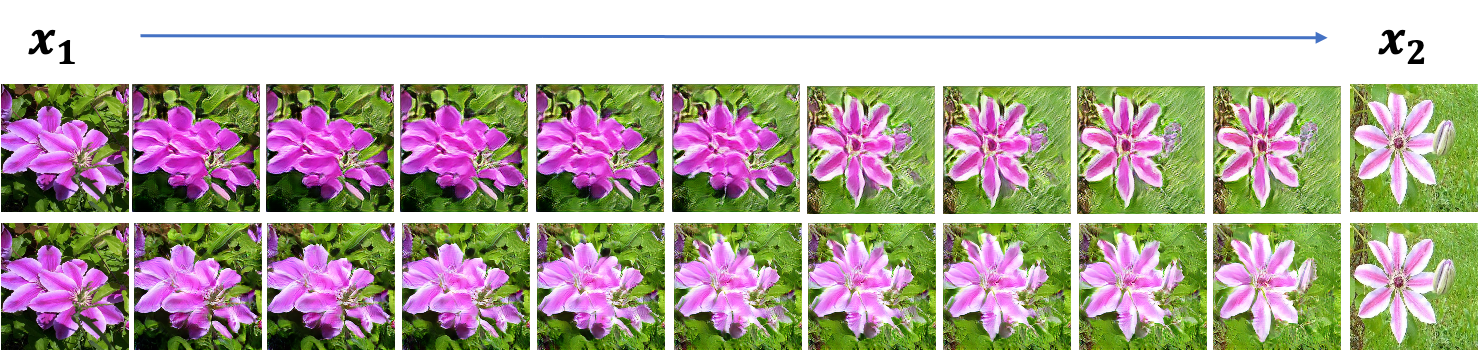}
\end{center}
\caption{Linear interpolation results of MatchingGAN (top row) and our F2GAN (bottom row) based on two conditional images $ \textbf{x}_1$ and $ \textbf{x}_2$ on Flowers dataset.}
\label{fig:interpolation} 
\end{figure}

\subsection{Few-shot Classification}
\label{sec:few-shot}

We follow the $N$-way $C$-shot setting in few-shot classification~\cite{vinyals2016matching,sung2018learning} by creating evaluation episodes and calculating the averaged accuracy over multiple evaluation episodes. In each evaluation episode, $N$ categories are randomly selected from unseen categories. Then, $C$ images from each of $N$ categories are randomly selected as training set while the remaining images are used as test set. We use pretrained ResNet$18$~\cite{he2016deep} (pretrained on the seen categories) as the feature extractor and train a linear classifier for the selected $N$ unseen categories. Besides $N\times C$ training images, our fusion generator produces $512$ additional images for each of $N$ categories to augment the training set. 

We compare our method with existing few-shot classification methods, including representative methods MatchingNets~\cite{vinyals2016matching}, RelationNets~\cite{sung2018learning}, MAML~\cite{finn2017model} as well as state-of-the-art methods MTL~\cite{sun2019meta}, DN4~\cite{li2019revisiting}, MatchingNet-LFT~\cite{Hungfewshot}. Note that no augmented images are added to the training set of $N$ unseen categories for these baseline methods. Instead, we strictly follow their original training procedure, in which the images from seen categories are used to train those few-shot classifiers. Among the baselines, MAML \cite{finn2017model} and MTL~\cite{sun2019meta} need to
further fine-tune the trained classifier based on the training set of $N$ unseen categories in each evaluation episode.

We also compare our method with competitive few-shot generation baseline MatchingGAN~\cite{hong2020matchinggan}. For MatchingGAN, We use the same setting as our F2GAN and generate augmented images for unseen categories. Besides, we compare our F2GAN with FUNIT~\cite{liu2019few} in Supplementary.

By taking $5$-way/$10$-way $5$-shot as examples, we report the averaged accuracy over $10$ episodes on three datasets in Table~\ref{tab:performance_fewshot_classifier}.
Our method achieves the best results in both settings on all datasets, which shows the benefit of using augmented images produced by our fusion generator.  

\begin{figure}
\begin{center}
\includegraphics[scale=0.47]{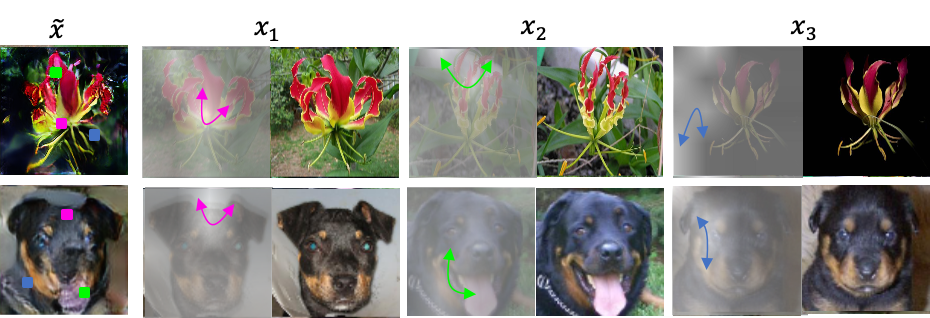}
\end{center}
\caption{The visualization of learned attention maps from NAF module. In each row, $\tilde{ \textbf{x}}$ is generated based on three conditional images $ \textbf{x}_1$, $ \textbf{x}_2$, and $ \textbf{x}_3$.
For each color dot (query location) in $\tilde{ \textbf{x}}$, we draw a color arrow with the same color to summarize the bright region (the most-attended region corresponding to the query location) in $ \textbf{x}_k$. Best viewed in color.}
\label{fig:attention_map} 
\end{figure}

\subsection{Ablation Studies}

\textbf{The number of conditional images} To analyze the impact of the number of conditional images, we train F2GAN with $K_1$ conditional images based on seen categories, and generate new images for unseen categories with $K_2$ conditional images. Due to space limitation, we leave the details to Supplementary.


      

\noindent\textbf{Loss terms: }In our method, we employ weighted reconstruction loss $\mathcal{L}_{1}$ (\ref{eqn:loss_reconstruction}), mode seeking loss $\mathcal{L}_{m}$ (\ref{eqn:loss_mode_seeking}), and interpolation regression loss $\mathcal{L}_{a}$ (\ref{eqn:loss_interpolation}). To investigate the impact of $\mathcal{L}_{1}$, $\mathcal{L}_{m}$, and $\mathcal{L}_{a}$, we conduct experiment on Flowers dataset by removing each loss term from the final objective (\ref{optimization}) separately. The quality of generated images is evaluated from two perspectives. On one hand, IS, FID, and LPIPS of generated images are computed as in Section~\ref{sec:visualization}. On the other hand, we report the accuracy of few-shot ($10$-way $5$-shot) classification augmented with generated images as in Section~\ref{sec:few-shot}. The results are summarized in Table~\ref{tab:network_design}, which shows that ablating $\mathcal{L}_1$ leads to slight degradation of generated images. Another observation is that without $\mathcal{L}_{m}$, the results \emph{w.r.t.} all metrics become much worse, which indicates that the mode seeking loss can enhance the diversity of generated images. Besides, when removing $\mathcal{L}_a$, it can be seen that the diversity and realism of generated images are compromised, resulting in lower classification accuracy.

\noindent\textbf{Attention module: }In our fusion generator, a Non-local Attentional Fusion (NAF) module is designed to borrow low-level information from the encoder. To corroborate the effectiveness of our design, we remove the NAF module, and directly connect the fused encoder features with the output of corresponding decoder blocks via skip connection, which is referred to as ``w/o NAF" in Table ~\ref{tab:network_design}. Besides, we replace our NAF module with local attention used in~\cite{lathuiliere2019attention} to compare two different attention mechanisms, which is referred to as ``local NAF" in Table~\ref{tab:network_design}. The results show that both ``local NAF" or our NAF achieve better results than ``w/o NAF", which proves the necessity of attention enhanced fusion strategy. We also observe that our NAF module can improve the realism and diversity of generated images, which is justified by lower FID, higher IS, and higher LPIPS. 

Moreover, we visualize the attention maps in Figure~\ref{fig:attention_map}. The first column exhibits the generated images based on three conditional images. For each generated image, we choose three representative query locations, which borrow low-level details from three conditional images respectively. For the conditional image $\textbf{x}_k$, 
we obtain the $H_1\times W_1$ attention map from the corresponding row in $\bm{A}_1^k$ in (\ref{eqn:attention_map}). For each color query point, we draw a color arrow with the same color to summarize the most-attended regions (bright regions) in the corresponding conditional image. In the first row, we can see that the red (\emph{resp.}, green, blue) query location in the generated flower $\tilde{\textbf{x}}$ borrows some color and shape details from $\textbf{x}_1$ (\emph{resp.}, $\textbf{x}_2$, $\textbf{x}_3$). Similarly, in the second row, the red (\emph{resp.}, green, blue) query location in the generated dog face $\tilde{\textbf{x}}$ borrows some visual details of forehead (\emph{resp.}, tongue, cheek) from $\textbf{x}_1$ (\emph{resp.}, $\textbf{x}_2$, $\textbf{x}_3$).

\setlength{\tabcolsep}{6pt}
\begin{table}[t]
  \caption{Ablation studies of our loss terms and attention module on Flowers dataset.} 
  \centering
  \begin{tabular}{lrrrr}  
    \hline
     setting& accuracy (\%)  & FID ($\downarrow$)  & IS ($\uparrow$) & LPIPS ($\uparrow$) \cr
    \hline
     w/o $\mathcal{L}_{1}$ & 74.89 &  122.68  & 6.39 & 0.2114  \cr
     w/o $\mathcal{L}_{m}$ & 73.92 & 125.26   & 4.92 & 0.1691\cr
     w/o $\mathcal{L}_{a}$ & 72.42 & 122.12    & 4.18 & 0.1463  \cr
     \hline
    w/o NAF & 72.62 & 137.81  & 5.11& 0.1825 \cr
    local NAF & 73.98 & 134.45 & 5.92 & 0.2052 \cr
    \hline
    F2GAN &$\textbf{75.02}$ &$\textbf{120.48}$ &$\textbf{6.58}$ &$\textbf{0.2172}$\cr
    \hline
  \end{tabular}
  \vspace{0.1mm}
  \label{tab:network_design}
\end{table}

\section{Conclusion}
In this paper, we have proposed a novel few-shot generation method F2GAN to fuse high-level features of conditional images and fill in the detailed information borrowed from conditional images. Technically, we have developed a non-local attentional fusion module and an interpolation regression loss. We have conducted extensive generation and classification experiments on five datasets to demonstrated the effectiveness of our method.

%
\begin{acks}
The work is supported by the National Key R$\&$D Program of China (2018AAA0100704) and is partially sponsored by National Natural Science Foundation of China (Grant No.61902247) and Shanghai Sailing Program (19YF1424400).
\end{acks}

%
\bibliographystyle{ACM-Reference-Format}
 \balance
\bibliography{f2gan}

\section{The number of conditional images} To analyze the impact of the number of conditional images, we train our F2GAN with $K_1$ conditional images based on seen categories, and generate new images for unseen categories with $K_2$ conditional images. By default, we set $K=K_1=K_2=3$ in our experiments. We evaluate the quality of generated images using different $K_1$ and $K_2$ in low-data (\emph{i.e.}, $10$-sample) classification (see Section $4.4$ in the main paper). By taking EMNIST dataset as an example, we report the results in Table~\ref{tab:number_effect} by varying $K_1$ and $K_2$ in the range of $[3, 5, 7, 9]$. From Table~\ref{tab:number_effect}, we can observe that our F2GAN can achieve satisfactory performance when $K_2=K_1$. The performance generally increases as $K$ increases (except from 3 to 5), but the performance gain is not very significant. Then, we observe the performance variance with fixed $K_1$.
Given a fixed $K_1$, when $K_2<K_1$, the performance drops a little compared with $K_2=K_1$. Instead, when $K_2>K_1$, the performance drops sharply, especially when $K_2$ is much larger than $K_1$ (\emph{e.g.}, $K_1=3$ and $K_2=9$). One possible explanation is that when we train our F2GAN with $K_1$ conditional images, it is not adept at fusing the information of more conditional images ($K_2>K_1$) in the testing phase.

\setlength{\tabcolsep}{8pt}
\begin{table}[t]
 \caption{Accuracy(\%) of low-data (10-sample) classification augmented by our F2GAN with different $ \textbf{K}_1$ and $ \textbf{K}_2$ on EMNIST dataset.} 
 \centering
  \fontsize{8}{8}\selectfont
 \begin{tabular}{lrrrr}
 \hline
    ~ & $K_1=3$ & $K_1=5$&$K_1=7$ & $K_1=9$  \cr
      
    \hline
    $K_2=3$&97.01  & 96.86  & 95.82 & 94.56  \cr
    \hline
    $K_2=5$& 95.24 &  96.98 & 96.08 & 95.52  \cr
    \hline
     $K_2=7$&93.76  &  95.13 & 97.23 & 96.86  \cr
    \hline
     $K_2=9$&90.17  & 92.74  &  94.38& 97.86  \cr
    \hline    
 \end{tabular}
  \vspace{0.1mm}
 \label{tab:number_effect}
\end{table}



\section{More Generation Results}
\begin{figure*}
\begin{center}
\includegraphics[scale=0.25]{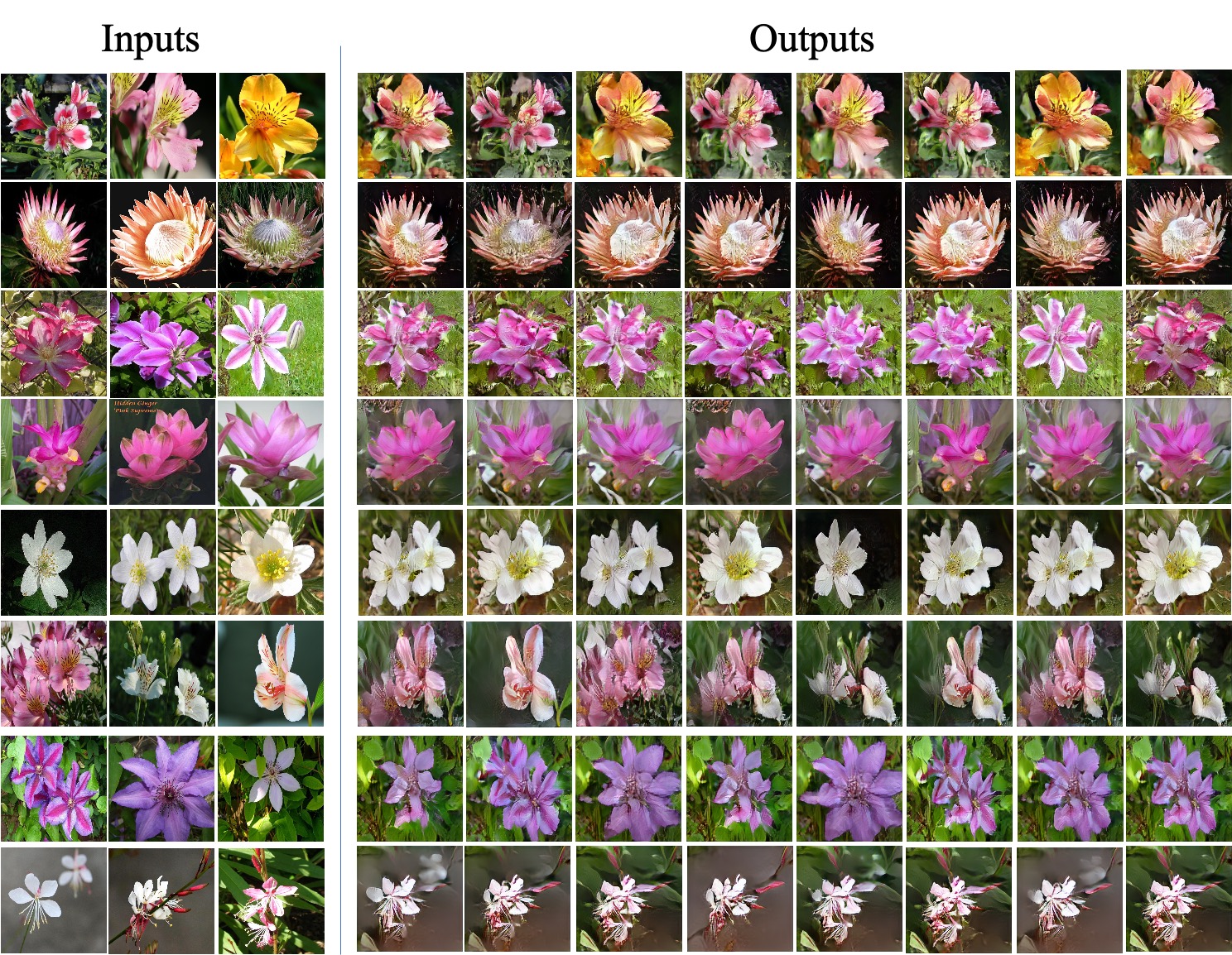}
\end{center}
\caption{Images generated by our F2GAN($ \textbf{K=3}$) on Flowers dataset. The conditional images are in the left three columns.} 
\label{fig:flowers} 
\end{figure*}

\begin{figure*}
\begin{center}
\includegraphics[scale=0.25]{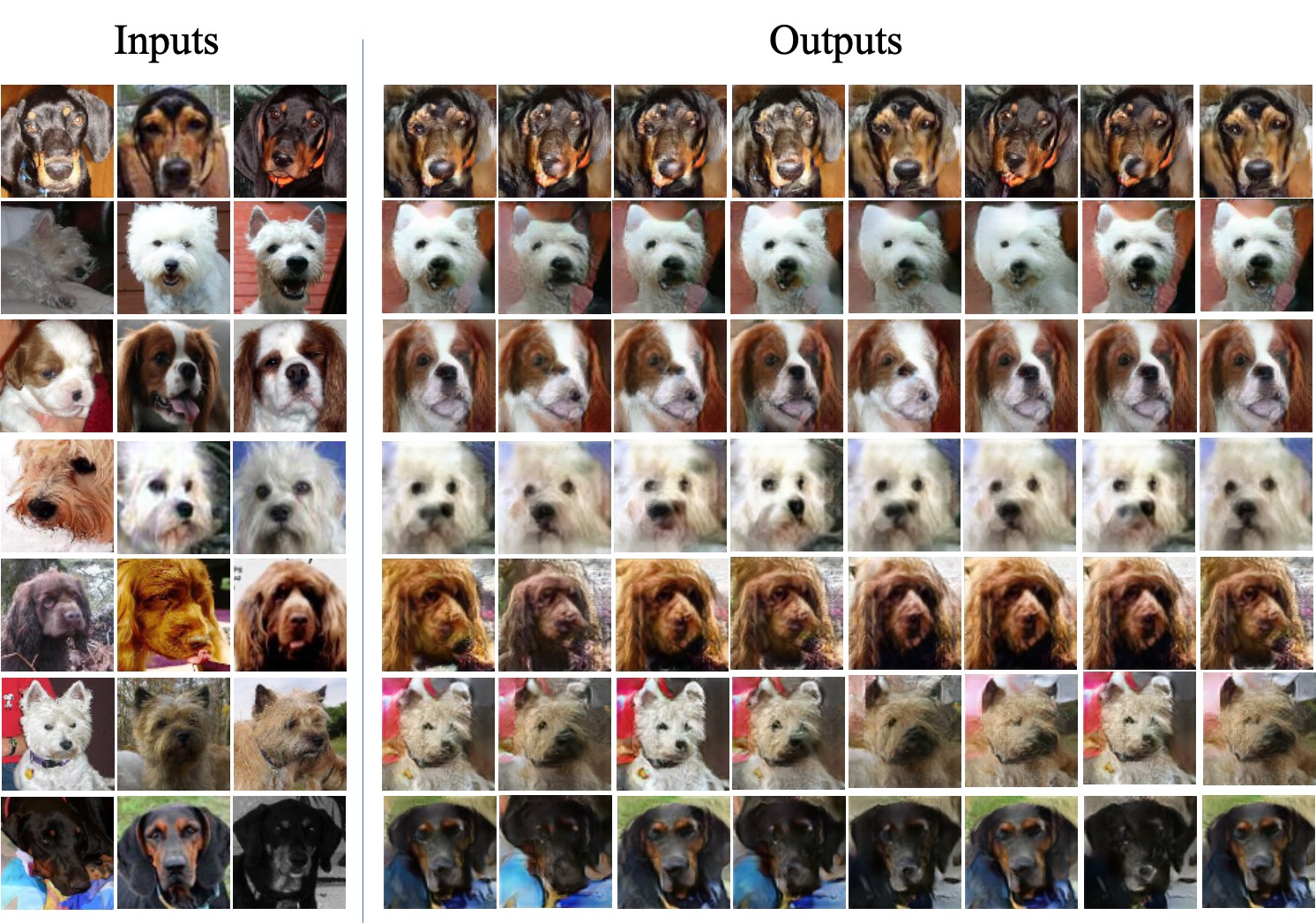}
\end{center}
\caption{Images generated by our F2GAN($ \textbf{K=3}$) on Animals Faces dataset. The conditional images are in the left three columns.} 
\label{fig:animals} 
\end{figure*}

\begin{figure*}
\begin{center}
\includegraphics[scale=0.5]{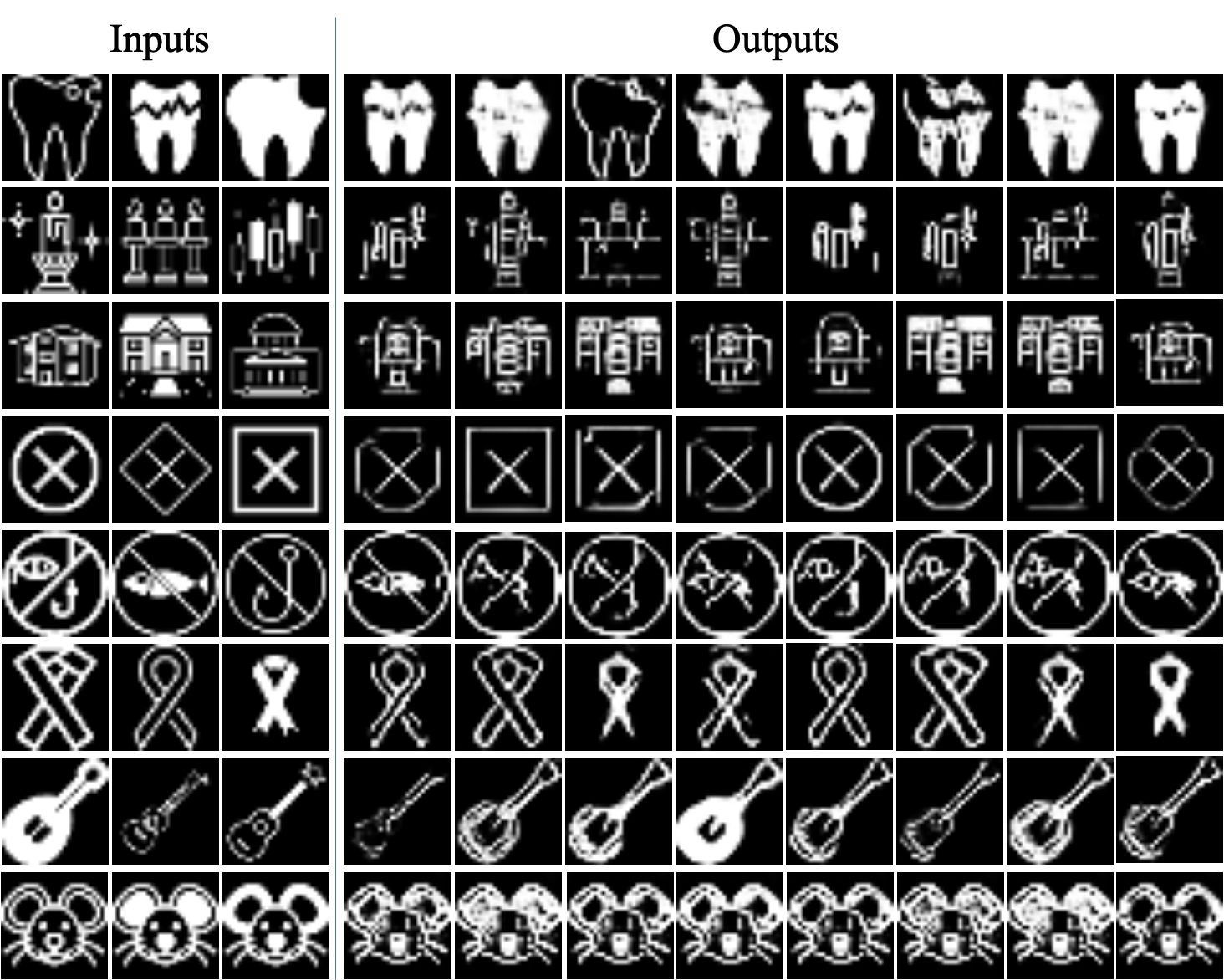}
\end{center}
\caption{Images generated by our F2GAN($ \textbf{K=3}$) on FIGR-8 dataset ~\cite{clouatre2019figr}. The conditional images are in the left three columns.} 
\label{fig:figr-8} 
\end{figure*}

We show more example images generated by our F2GAN ($K=3$) on Flowers and Animals datasets in Figure~\ref{fig:flowers} and Figure~\ref{fig:animals} respectively. Besides, we additionally conduct experiments on FIGR-8~\cite{clouatre2019figr} dataset, which is not used in our main paper. The generated images on FIGR-8 dataset are shown in Figure~\ref{fig:figr-8}. On all three datasets, our F2GAN can generally generate diverse and plausible images based on a few conditional images. However, for some complex categories with very large intra-class variance, the generated images are not very satisfactory. For example, in the $4$-th row in Figure~\ref{fig:animals}, the mouths of some dog faces look a little unnatural. We conjecture that in these hard cases, our fusion generator may have difficulty in fusing the high-level features of conditional images or seeking for relevant details from conditional images.

\section{More Interpolation results}
As in Section 4.3 in the main paper, We show more interpolation results of our F2GAN in Figure~\ref{fig:interpolation}. 
Given two images from the same unseen category, we perform linear interpolation based on these two conditional images. In detail, for interpolation coefficients $\bm{a}=[a^1, a^2]$, we start from $[0.9,0.1]$, and then gradually decrease (\emph{resp.}, increase) $a^1$ (\emph{resp.},  $a^2$) to $0.1$ (\emph{resp.}, $0.9$) with step size $0.1$.
It can be seen that our F2GAN is able to produce diverse and realistic images with rich details between two conditional images, even when two conditional images are quite different. 

\begin{figure*}
\begin{center}
\includegraphics[scale=0.6]{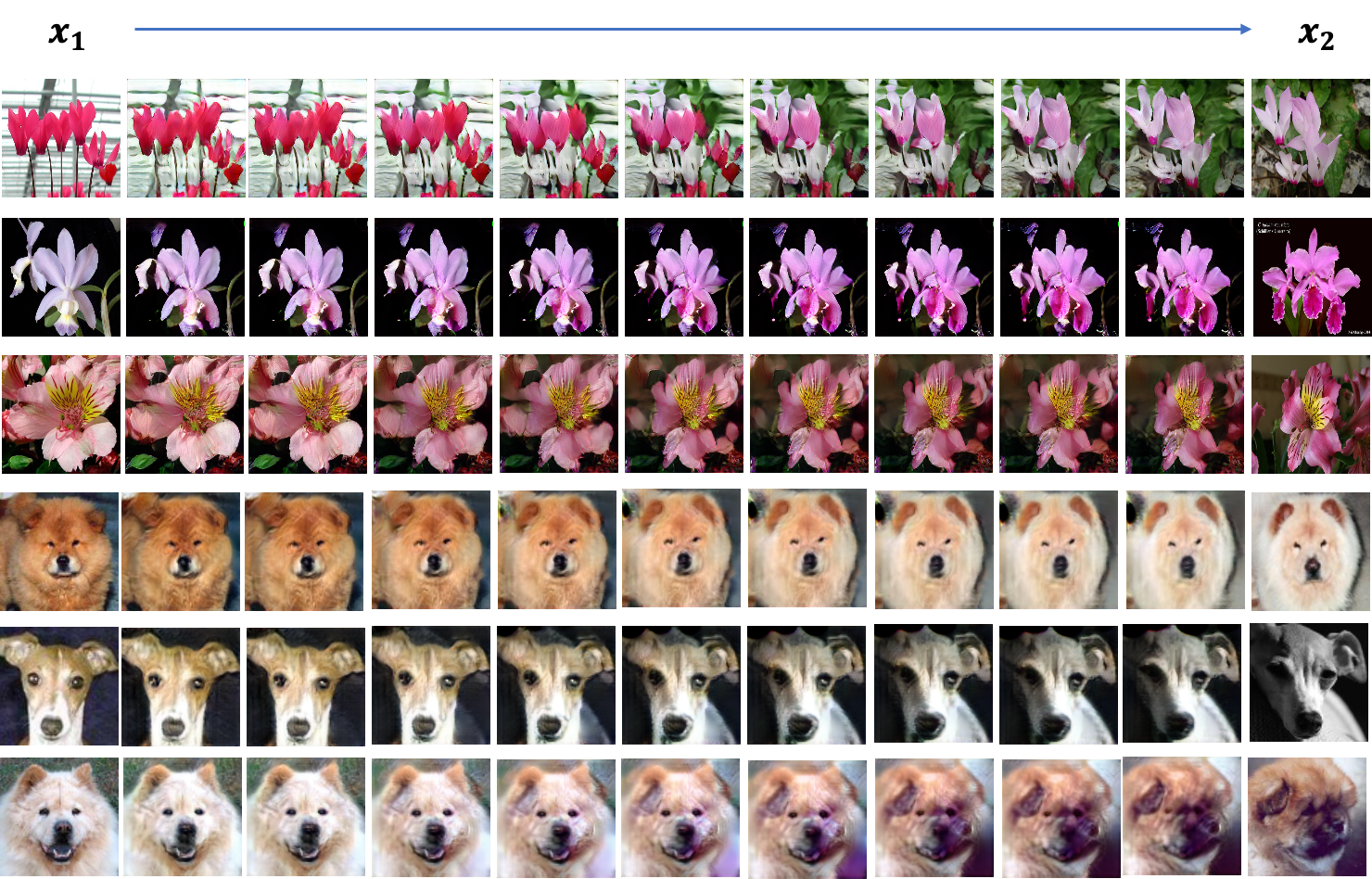}
\end{center}
\caption{Linear interpolation results of our F2GAN on Flowers dataset.}
\label{fig:interpolation} 
\end{figure*}


\section{Comparison with Few-shot Image Translation}
Few-shot image translation methods like FUNIT~\cite{liu2019few} mainly borrow category-invariant content information from seen categories to generate new images for unseen categories in the testing phase. 
Technically, FUNIT disentangles the category-relevant factors~(\emph{i.e.}, class code) and category-irrelevant factors~(\emph{i.e.}, content code) of images. Next, we refer to the images from seen (\emph{resp.}, unseen) categories as seen (\emph{resp.}, unseen) images.
By replacing the content code of an unseen image with those of seen images, FUNIT can generate more images for this unseen category.
However, in this way, few-shot image translation can only introduce category-irrelevant diversity, but fails to introduce enough category-relevant diversity for category-specific properties. 

To confirm this point, we conduct few-shot classification experiments (see Section $4.5$ in the main paper) to evaluate the quality of generated images. Based on the released model of FUNIT~\cite{liu2019few} trained on Animal Faces~\cite{deng2009imagenet}, we use class codes of unseen images and content codes of seen images to generate $512$ new images for each unseen category. Then, we use the generated images to help few-shot classification  (see Section $4.5$ in the main paper), which is referred to as ``FUNIT-1" in Table~\ref{tab:performance_fewshot_classifier}. Besides, we also exchange content codes within the images from the same unseen category to generate new images for each unseen category, but the number of new images generated in this way is quite limited. 
Specifically, in $N$-way $C$-shot setting, we can only generate $(C-1) \times C$ new images for each unseen category. We refer to this setting as ``FUNIT-2" in Table~\ref{tab:performance_fewshot_classifier}. 

From Table~\ref{tab:performance_fewshot_classifier}, it can be seen that ``FUNIT-1" is better than ``FUNIT-2", because ``FUNIT-1" leverages a large amount of extra seen images when generating new unseen images. However, ``FUNIT-1" is inferior to some state-of-the-art few-shot classification methods as well as our F2GAN, because FUNIT cannot introduce adequate category-relevant diversity as analyzed above.

\setlength{\tabcolsep}{2pt}
\begin{table}[t]
  \caption{Accuracy(\%) of different methods on Animals Faces in few-shot classification setting.} 
  \centering
  \begin{tabular}{lrr}
      \hline
      Method & 5-way 5-shot &10-way 5-shot\cr
    \hline
    MatchingNets~\cite{vinyals2016matching}  &59.12  &50.12 \cr

    MAML~\cite{finn2017model} & 60.03  &49.89 \cr

    RelationNets~\cite{sung2018learning} &67.51  & 58.12 \cr

    MTL~\cite{sun2019meta} &79.85  &70.91 \cr

    DN4~\cite{li2019revisiting} &81.13  &71.34 \cr
    
    MatchingNet-LFT~\cite{Hungfewshot}  &80.95  &71.62 \cr
    MatchingGAN~\cite{hong2020matchinggan} & 80.36 & 70.89\cr
    FUNIT-1 & 78.02 &69.12  \cr
    FUNIT-2 &75.29  &67.87  \cr
    F2GAN & $\textbf{82.69}$ & $\textbf{73.19}$ \cr
    \bottomrule[0.8pt]
    \end{tabular}
  \label{tab:performance_fewshot_classifier}
\end{table}

\section{Details of Network Architecture}
\textbf{Generator} In our fusion generator, there are in total $11$  residual blocks ($5$ encoder blocks, $5$ decoder blocks, and $1$ intermediate block), in which each encoder (\emph{resp.},decoder) block contains $3$ convolutional layers with leaky ReLU and batch normalization followed by one downsampling (\emph{resp.}, upsampling) layer, while intermediate block contains $3$ convolutional layers with leaky ReLU and batch normalization. The architecture of our generator is summarized in Table~\ref{tab:generator}. 

\setlength{\tabcolsep}{8pt}
\begin{table}[t]
  \caption{The network architecture of our fusion generator. BN denotes batch normalization.} 
  \centering
  \begin{tabular}{ccccc}
  \hline
   Layer & Resample & Norm & Output Shape  \cr
      
    \hline
     Image $\bm{x}$  & -   &  - & 128*128*3   \cr
     \hline
     Conv $1 \times 1$  &  -  &  - &  128*128*32  \cr
     \hline
     Residual Block  &  AvgPool  & BN  &  64*64*64  \cr
     \hline
     Residual Block  &   AvgPool  & BN  &  32*32*64  \cr
     \hline
     Residual Block  &   AvgPool  & BN &  16*16*96  \cr
     \hline
     Residual Block  &   AvgPool  & BN &  8*8*96  \cr
     \hline
     Residual Block  &   AvgPool  & BN &  4*4*128  \cr
     \hline
      Residual Block  &   -  & BN &  4*4*128  \cr
     \hline
     Residual Block  &   Upsample  & BN  &   8*8*96 \cr
     \hline
     Residual Block  &  Upsample  & BN  &  16*16*96  \cr
     \hline
     Residual Block  &  Upsample  & BN &   32*32*64 \cr
     \hline
     Residual Block  &  Upsample  & BN  & 64*64*64   \cr
     \hline
     Residual Block  &  Upsample  & BN  & 128*128*64   \cr
     \hline
     Conv $1 \times 1$  & - & - &   128*128*3 \cr
    \hline    
  \end{tabular}
  \label{tab:generator}
\end{table}

\noindent\textbf{Discriminator} Our discriminator is analogous to that in~\cite{liu2019few}, which consists of one convolutional layer followed by five groups of residual blocks. Each group of residual blocks is as follows: ResBlk-$k$ $\rightarrow$ ResBlk-$k$ $\rightarrow$ AvePool$2$x$2$, where ResBlk-$k$ is a ReLU first residual block~\cite{mescheder2018training} with the number of channels $k$ set as $64$, $128$, $256$, $512$, $1024$ in five residual blocks. We use one fully connected (fc) layer with $1$ output following global average pooling layer to obtain the discriminator score. The architecture of our discriminator is summarized in Table ~\ref{tab:discriminator}. 

The classifier shares the feature extractor with the discriminator and only replaces the last fc layer with another fc layer with $C^s$ outputs with $C^s$ being the number of seen categories. The mode seeking loss and the interpolation regression loss also use the feature extractor from the discriminator. Specifically, we remove the last fc layer from discriminator to extract the  features of generated images, based on which the mode seeking loss and the interpolation regression loss are calculated.

\setlength{\tabcolsep}{8pt}
\begin{table}[t]
  \caption{The network architecture of our fusion discriminator.} 
  \centering
  \begin{tabular}{ccccc}
  \hline
    Layer & Resample & Norm & Output Shape  \cr
      
    \hline
     Image $\bm{x}$  & -   &  - & 128*128*3   \cr
     \hline
     Conv $1 \times 1$  &  -  &  - &  128*128*32  \cr
     \hline
      Residual Blocks  &  AvgPool  & -  &  64*64*64  \cr
     \hline
      Residual Blocks  &   AvgPool  & -  &  32*32*128  \cr
     \hline
      Residual Blocks  &   AvgPool  & - &  16*16*256  \cr
     \hline
     Residual Blocks  &   AvgPool  & - &  8*8*512  \cr
     \hline
     Residual Blocks  &   AvgPool  & - &  4*4*1024  \cr
     \hline
     Global  &   GlobalAvgPool  & - &  1*1*1024  \cr
      \hline
      FC  &   -  & - &  1*1*1  \cr
      \hline
  \end{tabular}
  \label{tab:discriminator}
\end{table}

\end{document}